\documentclass[letterpaper, 10 pt, conference]{ieeeconf}  

\IEEEoverridecommandlockouts                              
\usepackage{graphicx}
\usepackage{amsmath}
\usepackage{amssymb}
\usepackage[table]{xcolor}
\usepackage{algorithm,algorithmic}

\newcommand{\cc}{\cellcolor{green!25}}

\renewcommand{\mathbf}{\boldsymbol}

\title{\LARGE \bf
Randomised Algorithm for Feature Selection and Classification}

\author{Aida~Brankovic, Alessandro~Falsone, Maria~Prandini, Luigi~Piroddi
\thanks{Dipartimento di Elettronica, Informazione e Bioingegneria, Politecnico di Milano, Italy.
        {\tt\small \{aida.brankovic, alessandro.falsone, luigi.piroddi\} at polimi.it  }}%
}

\begin{document}

\newtheorem{lemma}{Lemma}[section]
\newtheorem{theorem}{Theorem}[section]
\newtheorem{corollary}{Corollary}[section]
\newtheorem{remark}{Remark}[section]
\newtheorem{proposition}{Proposition}[section]
\newtheorem{definition}{Definition}[section]

\newenvironment{Proof}{{\noindent \bf Proof\ }}{}

\maketitle
\thispagestyle{empty}
\pagestyle{empty}

\begin{abstract}
We here introduce a novel classification approach adopted from the nonlinear model identification framework, which jointly addresses the feature selection and classifier design tasks. The classifier is constructed as a polynomial expansion of the original attributes 
and a model structure selection process is applied to find the relevant terms of the model. The selection method progressively refines a probability distribution defined on the model structure space, by extracting sample models from the current distribution and using the aggregate information obtained from the evaluation of the population of models to reinforce the probability of extracting the most important terms. To reduce the initial search space, distance correlation filtering can be applied as a preprocessing technique.
The proposed method is evaluated and compared to other well-known feature selection and classification methods on standard benchmark classification problems. The results show the effectiveness of the proposed method with respect to competitor methods both in terms of classification accuracy and model complexity. The obtained models have a simple structure, easily amenable to interpretation and analysis.

\begin{keywords}
Feature selection, Classification, Nonlinear identification, Model selection, Randomized methods
\end{keywords}
\end{abstract}

\section{Introduction}
\label{sec:Intro}

In the supervised learning framework, classification is the task of predicting the class labels of unseen observations (each consisting of a set of measured attributes or features), based on the experience gathered through a learning process on a previously available set of observations whose classes are known (training set). The classification task is generally decomposed into two stages, namely a first preprocessing stage denoted feature selection (FS), followed by the actual classifier design. FS is a combinatorial optimization problem which aims at extracting the relevant features from a given set. An effective FS procedure greatly facilitates the classifier design process, reducing its computational demand, simplifying the classifier structure, and ultimately improving the classification performance, which may be adversely affected by irrelevant and redundant features \cite{dash1997feature}. FS is particularly crucial and hard in problems with a large number of features, resulting in a huge search space (``curse of dimensionality'').

FS methods are often classified according to how strongly the attribute selection and model construction processes interact. In filter methods, FS is performed independently of the classifier design, based only on intrinsic properties of the features, whereas in wrapper methods a subset of the features is evaluated based on the classification performance it can achieve if it is used to build the classifier. The classifier operates on the selected features processing them through a linear or nonlinear model. In this process it may be useful to derive additional more complex features as functions of the original ones (feature extraction).

Filter methods (see \emph{e.g.}, \cite{chandrashekar2014survey}) can be used effectively to eliminate the least important terms, but they cannot be fully relied upon to eliminate all redundant terms since they do not take into account the interaction between features. Such relationships between regressors may indeed have an important impact on the selection process. For example, individually important features may become redundant when considered in combination with others, and individually irrelevant or redundant features may become relevant in combination with others \cite{xue2014particle}.

Wrapper methods are typically more accurate, though they are computationally intensive and may suffer from overfitting problems \cite{liu2012feature}. Several wrapper algorithms based on sequential search have been discussed in the literature, such as the PTA$(l,r)$ (Plus $l$ and Take Away $r$), the SFFS (Sequential Forward Floating Selection algorithm) and the SBFS (Sequential Backward Floating Selection algorithm) \cite{ferri1994comparative}. In these schemes, the algorithm starts from either the empty or the full set of features, and then features are iteratively added or removed. Similar incremental model building schemes have been developed in the context of nonlinear identification, particularly with reference to polynomial NARX/NARMAX models \cite{korenberg1988orthogonal}, \cite{piroddi2003identification}, \cite{Billings13}. Besides sequential approaches, a significant amount of work has been devoted to evolutionary methods such as Genetic Algorithms (GA) \cite{smith2005genetic}, \cite{yang1998feature}, Particle Swarm Optimization (PSO) \cite{xue2014particle}, \cite{xue2013particle}, Ant Colony Optimization (ACO) \cite{kabir2012new}, Harmony Search (HS) \cite{diao2012feature}.

Regarding the classifier design problem, several algorithms have been proposed in the literature, based on Artificial Neural Networks (ANN) \cite{paliwal2009neural}, Support Vector Machines (SVMs), often in combination with Radial Basis Functions (RBF) as kernel functions \cite{gunn1998support}, instance based learning methods such as Nearest Neighbor (NN) and Data Gravitational Classification (DGC) \cite{aha1991instance}, evolutionary methods as GP \cite{espejo2010survey} and PSO \cite{lin2008particle}.
Instance based algorithms are particularly appealing due to their simple classification logic and generally satisfactory performance. In the NN (or 1-NN) algorithm a new sample is simply assigned to the class of the nearest previously available sample. This is one of the most used and well known classification algorithms due its simplicity, though it is reported to suffer from various drawbacks such as high dimensionality, low efficiency, high storage requirements and sensitivity to noise \cite{triguero2011differential}. Cases where the classes are nonseparable or overlapping appear to be particularly critical \cite{li2008nearest}. The $k$-NN extension classifies a new instance based on the majority of the $k$ nearest neighbors. Several variants of the $k$-NN method have been proposed in the literature, that typically introduce weighted distances or similar concepts to improve the performance, such as the KNN-A \cite{wang2007improving}, the DW-KNN \cite{dudani1976distance}, the CNN \cite{gao2007center}, the CamNN \cite{zhou2006improving}. 
DGC algorithms \cite{cano2013weighted} have also been put forward as an attempt to overcome the mentioned problems of the NN algorithm. In this respect, it is worth mentioning the work of Peng \emph{et al.} \cite{peng2009data}, which employs feature weighting and a tentative random feature selection algorithm to compute the feature weights. An enhancement of the DCG algorithm, denoted DGC+ is proposed in \cite{cano2013weighted} to deal with imbalanced data.

Feature selection is also employed in different contexts from classification. For example, it is a crucial task in the identification of nonlinear dynamical models. In that framework, the model output is not a discrete variable, but rather a continuous one, and the input-output mapping typically configures a dynamical model. However, the model is still interpretable as a (nonlinear) mapping between a set of features (the lagged input and output samples) and the current output. The model is typically parameterized as a \emph{linear} combination of nonlinear terms derived from the features, \emph{e.g.} monomials. Such recursive input-output models are denoted NARX/NARMAX models, depending on how the noise entering the model is described 
\cite{Billings13}. The selection of the most appropriate of these extended nonlinear features (denoted model structure selection or briefly MSS) constitutes the main challenge in nonlinear identification. An important difference with the classification-oriented approach described previously is that rather than constructing a \emph{nonlinear} mapping from features to classes, the feature set is first extended to include the mentioned nonlinear features, and then a \emph{linear} map from these to the output is constructed. This is particularly convenient in that the model configures a linear regression, whose parameters can be estimated by means of well-assessed techniques. In view of this, the extended nonlinear features are often called regressors.

A pioneer method in the nonlinear identification framework is the Forward Regression Orthogonal Estimator (FROE) \cite{korenberg1988orthogonal}, which is based on an incremental building procedure, akin to the sequential wrapper approaches mentioned previously. At every iteration one regressor is added to the model, based on an importance rating. The latter is essentially a measure of the accuracy improvement (in terms of output prediction) that can be achieved by adding that particular term to the current model. 
A limitation of the FROE is that the importance rating provides only a relative measure of the importance of a regressor, which varies considerably depending on the other terms included in the model. Indeed, individually important regressors may become redundant in combination with others, while terms of scarce individual importance may become highly relevant in combination with others. This fact alone may greatly affect the correctness of the MSS process.

More reliable results can be obtained if the importance of the regressors is assessed based on information gathered from a \emph{population} of models, as is done \emph{e.g.} in the RaMSS (Randomized MSS) algorithm \cite{Falsone2015227}. Accordingly, we here propose a wrapper algorithm that is a modified version of the RaMSS adapted for classification purposes, denoted in the following RFSC (Randomized Feature Selection and Classifier). In accordance with the MSS philosophy, first a set of extended nonlinear features is generated from the original ones, using polynomial expansions, and the classification model is defined as a linear combination of a subset of these extended features. A probability distribution is defined over the model structure space, that describes the probability of each possible extended feature subset to be the true model structure. At each iteration, a population of sample model structures is extracted from the current distribution and all the corresponding models are estimated and evaluated. Parameter estimation is facilitated by the linear-in-the-parameters structure of the classification models.

Then, the probability distribution is updated, by reinforcing the probability to extract those regressors that appear in accurate models more often than not, and accordingly reducing the probability to extract the remaining ones. Note that in doing so the importance of a regressor is not estimated any more based on a local measure in the model space, but rather on the aggregate information associated to an entire population of models. The method progressively refines the probability distribution until it converges to a limit distribution associated to a single model. Experimental results show that this method 
provides quite compact and accurate models. The RFSC algorithm does not suffer from error accumulation problems that may be observed with sequential methods, and generally bases the selection of features on more robust evidence than what may be gathered from individual models. Also, the randomization inherent in the approach yields sufficient exploration capabilities to allow the algorithm to occasionally escape from local minima.

The search space rapidly increases with the number of features and the order of the polynomial expansion. Large-sized search spaces complicate the FS task and may adversely affect the search process. To address this issue, a dependency test based on the \emph{correlation of distances} \cite{szekely2007} has been carried out for medium/large size problems. Distance correlation provides a reliable dependency measure between random vectors, and can be used to test the individual dependence of the output vector on each feature vector. Only features with enough statistical evidence to reject the independence hypothesis with a given significance level are considered in the FS process, thus reducing the search space.



The rest of the paper is organized as follows. Section \ref{sec:Prelim} presents the classification problem in a nonlinear regression framework. A probabilistic formulation of the feature selection problem is then proposed in Section \ref{sec:ProbForm}. This constitutes the basis for the development of the RFSC method, described in Section \ref{sec:AlgDecr}. Several numerical studies on benchmark datasets are discussed in Section \ref{sec:ExpStdy}. 
Finally, Section \ref{sec:concl} presents some concluding remarks.

\section{Preliminaries}
\label{sec:Prelim}

\subsection{The classification problem}


In classification problems one is interested in constructing a model that captures the relationship between features (inputs) and classes (outputs) through a learning process operating on available observations (input-output pairs). Classification is akin to model identification, the main differences being that the input-output relationship is typically non-dynamic and that the outputs (and sometimes the inputs) take values in a discrete set. This similarity makes it sometimes possible to adapt algorithms developed in the identification domain to solve classification tasks, as endeavored here.

Assume that a set of $N$ observations is available, each consisting of a pair $(\mathbf{u}(k),c(k))$, $k = 1, \ldots, N$, where the components $u_p$, $p = 1, \ldots, N_f$ of vector $\mathbf{u}$ are the features and $c \in \{ 1, \ldots, N_c \}$ is the corresponding class (assumed known, according to the supervised learning framework). In the following, we adopt a one-vs.-rest strategy to deal with multi-class problems, and accordingly recode the output as an $N_c$-dimensional vector $\mathbf{y}$, with binary components, defined as:
\begin{align}
\label{eq:OutputCoding}
&y_i(k) = \begin{cases}
1, & c(k) = i\\
-1, & \text{otherwise}
\end{cases}
\end{align}
where $i=1, \ldots, N_c$. Notice that if $N_c=2$, a single output is sufficient to discriminate between the two classes, the $-1$ value of $y_1$ being directly associated to class $2$.

The objective is to construct a classification model of the type:
\begin{equation}
\label{eq:feat_probl}
   \hat{\mathbf{y}}(k)=\mathbf{f}(\mathbf{u}(k)),
\end{equation}
where $\hat{\mathbf{y}}$ denotes the class estimate and $\mathbf{f}$ is a vector of unknown functions, that is capable of predicting correctly the class for observations unseen in the learning phase. Following the one-vs.-rest strategy, a separate model is devoted to the assessment of each class.
To avoid ambiguities in the class estimation, the actual class estimate is conventionally assumed as the label corresponding to the individual classifier returning the largest value:
\begin{equation*}
  \hat{c}(k) = \arg \underset{i=1, \ldots, N_c}{\max} \hat{y}_i(k)
\end{equation*}

In view of this, the multi-class problem can be addressed by training one binary classifier for each class, that discriminates if a sample belongs to one class or not. Accordingly, in the following we shall focus on the training and evaluation of the binary classifiers $\hat{y}_i(k)$, $i=1, \ldots, N_c$.

Regarding the unknown functions $f_i(\cdot)$, $i=1, \ldots, N_c$, a common approach is to represent them using parametric functional expansions
, so that:
\begin{equation}
  \label{eq:nl_reg}
  \hat{y}_i(k) = \left(\sum_{j=1}^{N_r} \vartheta_j^{(i)} \varphi_j(k) \right)
         = \Phi(k)^T \vartheta^{(i)},
\end{equation}
$i=1, \ldots, N_c$, where $\vartheta^{(i)}$ is a vector of unknown parameters (associated to the $i$th output), and $\Phi(k) = [\varphi_1(k) \ldots \varphi_{N_r}(k) ]^T$, where $\varphi_j(k) = \varphi_j(\mathbf{u}(k))$, $j = 1, \ldots, N_r$, is a given nonlinear function of the features. In view of the fact that equation \eqref{eq:nl_reg} actually configures a linear regression, these extended features are also called regressors.

Various types of basis functions have been used to construct the functional expansions, such as Fourier series, piecewise linear models, polynomial models, radial basis functions, and sigmoidal neural networks, all having the universal approximation property. In this work, we will consider polynomial expansions, so that the generic term $\varphi_j(k)$ takes the form:
\begin{align}
  \label{eq:regressor}
  &\varphi_j(k) = \begin{cases}
  u_{p_1}(k) \cdot u_{p_2}(k) \cdot \ldots \cdot u_{p_l}(k), & l > 0\\
  1, & l=0
  \end{cases}
\end{align}
where $p_s \in \{1, \ldots, N_f \}$, $s = 1, \ldots, l$, with $0 \leq l \leq M$, $M$ being the maximum allowed degree of the polynomial expansion.

This formulation has the advantage that 
the model is linear-in-the-parameters, which greatly facilitates parameter estimation. On the other hand, the number of terms in a polynomial expansion increases rapidly with the maximum degree and the number of arguments (curse of dimensionality). Conventional practice has it that relatively small models of this category are suitable for various applications. It is also well-known that the smaller the size of the model, the more robust it will be and the more capable of generalizing to new observations. Therefore, a crucial problem consists in selecting the best terms of type \eqref{eq:regressor} for the model, a task which is equivalent to feature selection, but applied to an extended set of features (constructed as monomials of the original ones).

\subsection{Parameter estimation of the $i$th component of the classifier}
\label{subsec:par_est}

As already mentioned the modeling task is addressed separately for each class. In the following we shall focus on the modeling of classifier $\hat{y}_i$ associated to class $i$. For ease of notation we will drop the indexing on class $i$.

For a given structure, the parameter estimation for a model of type \eqref{eq:nl_reg} is typically formulated as a minimization problem with reference to a loss function defined as $\mathcal{L} :\{-1,+1\} \times \mathbb{R} \to \mathbb{R}_+$. A standard loss function evaluates the model performance as the percentage of misclassified observations (with respect to class $i$). The resulting optimization problem is given by
\begin{equation}
\label{eq:minprob}
    \min_{\vartheta} \; \frac{1}{N} \sum_{k=1}^N \mathcal{L}_{0-1}(y(k),\hat{y}(k)),
\end{equation}
where $\mathcal{L}_{0-1}$ is the \emph{0-1 loss function}, defined as $\mathcal{L}_{0-1}(z_1,z_2)= \mathbf{1}_{\{z_1z_2 < 0\}}(z_1,z_2)$. Due to the hard nonlinearity enforced by this loss function, the latter is usually approximated with functions with nicer properties regarding optimization, such as the hinge loss, the squared hinge loss, the logistic loss, the exponential loss. In the following, the \emph{logistic loss} will be employed for this purpose, resulting in the following reformulation of the optimization problem:
\begin{equation}
   \label{eq:minprobnew}
   \min_{\vartheta} \; \frac{1}{N} \sum_{k=1}^N \log(1+e^{-y(k) \hat{y}(k)}).
\end{equation}
The reader should note that $\log(1+e^{(\cdot)})$ is a strictly convex function, and $\hat{y}(k)$ is linear in $\vartheta$. Therefore, the resulting cost function is strictly convex in $\vartheta$, and the minimizer of \eqref{eq:minprobnew} is unique.


Although there does not exist a closed-form solution to the above optimization problem, the logistic loss is a continuous differentiable function, which allows to apply gradient descent methods in the optimization process. 
In this work a standard Newton's iterative optimization scheme is applied to solve problem \eqref{eq:minprobnew}.

\subsection{Statistical test for regressor significance}
The rejection of redundant terms is a crucial step in the identification procedure. For this purpose, a statistical test is carried out after the parameter estimation phase to rule out terms whose parameters are statistically indistinguishable from $0$. According to \cite{bishop2006}, the update equation of the Newton method is structurally equivalent to an Iteratively Reweighted Least Squares algorithm, so that upon convergence one can evaluate the parameter covariance as in standard Weighted Least Squares.
Therefore, the variance $\hat{\sigma}^2_j$ of the estimated parameters is given by:
\begin{equation}
\label{eq:varest}
  \hat{\sigma}^2_j \approx \hat{\sigma}^2_e G_{jj}^{-1},
\end{equation}
where $\hat{\sigma}^2_e$ is variance of the residuals and $G_{jj}$ is the $j$th diagonal element of the Hessian $G = {\Psi}^T R \Psi$ upon convergence, $\Psi = [\Phi(1) \ldots \Phi(N)]^T$ being a matrix containing all samples of the selected nonlinear regressors and $R$ a diagonal $N \times N$ matrix with diagonal elements given by:
\begin{equation*}
R_{kk} = \tilde{y}(k)(1-\tilde{y}(k)),
\end{equation*}
$k = 1, \ldots, N$, where $\tilde{y}(k) = 1/(1+e^{y(k) \hat{y}(k)})$.

The variance \eqref{eq:varest} can be employed in a Student's $t$-test to determine the statistical relevance of each regressor \cite{Falsone2015227}. More precisely, the $j$th regressor is considered to be statistically irrelevant (and therefore rejected) if the interval
\begin{equation}
   [\hat{\vartheta}_j-\hat{\sigma}_j t_{\alpha,N-\tau}, \hat{\vartheta}_j+\hat{\sigma}_j t_{\alpha,N-\tau}]
   \label{eq:stat_tets}
\end{equation}
contains $0$, $t_{\alpha,N-\tau}$ being the $100(1-\alpha)$ percentile of the Student's $t$ distribution with $N-\tau$ degrees of freedom and significance confidence interval $\alpha$, where $\tau$ is the number of components of $\vartheta$.

\section{Probabilistic formulation of the FS problem}
\label{sec:ProbForm}
This section addresses the FS problem for the $i$th component of the classifier. As done in the previous section, we drop the indexing on class $i$, for simplicity.

FS can be envisaged as the problem of finding the subset of regressors that maximizes the performance index $J$:
\begin{equation}
	J(f)=1-\frac{1}{N} \sum_{k=1}^N \mathcal{L}_{0-1}(y(k),\hat{y}(f;k))
\label{eq:perfIndex}
\end{equation}
over all possible model structures $f$ in the set $\mathcal{F} = 2^\mathcal{R}$, $\mathcal{R} = \{ \varphi_1, \cdots, \varphi_{N_r} \}$ being the full set of $N_r$ regressors. Notice that the dependence of the $\hat{y}$ on the model structure $f$ has been explicitly stated in Equation~\eqref{eq:perfIndex}.

This problem is combinatorial, in that in principle one would need to explore all $2^{N_r}$ model structures. A convenient solution approach involves a reformulation in a probabilistic framework \cite{Falsone2015227} by introducing the random variable $\phi$ which takes values in the set of all possible models $\mathcal{F}$ according to a probability distribution $\mathcal{P}_\phi$. The performance of $\phi$ is also a random variable, and its expectation is given by
\begin{equation} \label{eq:probForm0}
     \mathbb{E}[J(\phi)]= \sum_{f \in \mathcal{F}} J(f)\mathcal{P}_{\phi}(f).
\end{equation}
Index \eqref{eq:probForm0} is maximized when the probability mass concentrates on the model structure associated to the highest value of $J$ (or one of the possible best model structures, if the minimum is not unique). 
Therefore, the problem of finding the best $f \in \mathcal{F}$ can be formulated as
\begin{equation}
\label{eq:probForm}
  \mathcal{P}_{\phi}^* =  \arg \underset{\mathcal{P}_{\phi}} {\max \;} \mathbb{E}[J(\phi)],
\end{equation}
where $\mathcal{P}_{\phi}^*$ is such that $\mathcal{P}^*_{\phi}(f^*) = 1$.

A convenient parametrization for $ \mathcal{P}_{\phi}$ is obtained by associating a Bernoulli random variable $\rho_j$ to each regressor $\varphi_j$, that models the probability that $\varphi_j$ belongs to the target model:
\begin{equation*}
\label{eq:Bnl}
  \rho_j \sim Be(\mu_j),
\end{equation*}
$j=1, \ldots, N_r$, where $\mu_j \in [0,1]$. According to this representation, a model extraction from $ \mathcal{P}_\phi$ implies testing each regressor for inclusion, by extracting a value from the respective Bernoullian distribution. Regressor $\varphi_j$ is included if the outcome of the $j$th extraction is $1$, and omitted in case of $0$. The former event has probability $\mu_j$, whereas the probability of getting a $0$ is given by $1-\mu_j$. Accordingly, in the rest of the paper we will denote $\mu_j$ as the \emph{Regressor Inclusion Probability} (RIP) of the $j$th regressor, and define $\mu = [\mu_1 \cdots \mu_{N_r}]^T$ as the vector of RIPs. For simplicity, we assume that all random variables $\rho_j$, $j=1, \ldots, N_r$ are independent. In summary, the probability distribution $\mathcal{P}_\phi$ over the models in $\mathcal{F}$ can be written as:
\begin{equation}
\label{eq:Pphi}
   \mathcal{P}_\phi(f)= \prod_{j: \varphi_j \in f} \mu_j \prod_{j: \varphi_j \notin f} (1-\mu_j)
\end{equation}
for any $f \in \mathcal{F}$. If all RIPs have values in $\{0, 1\}$ only, a limit distribution is obtained with all probability mass concentrated on a specific model $\tilde{f}$ (containing all the regressors whose RIPs equal $1$). In that case, it follows that $\mathcal{P}_\phi(\tilde{f})=1$. The objective of the FS procedure will therefore be that of adapting the RIPs until convergence to the target limit distribution associated to an optimal model $f^*$.

%
%
%
%

To evaluate the importance of a given term we consider an aggregate indicator $\mathcal{I}_j$ that compares the average performance of the models including the $j$th regressor with that of the remaining ones:
\begin{equation}
\label{eq:Ii}
   \mathcal{I}_j= \mathbb{E} [J(\phi)| \varphi_j \in \phi] - \mathbb{E} [J(\phi)| \varphi_j \notin \phi],
\end{equation}
where $j=1, \dots, N_r$. 
The interested reader is referred to \cite{Falsone2015227} for all the mathematical details.
Thanks to the averaging over all models, indicator $\mathcal{I}_j$ can be interpreted as a global measure of the regressor importance. In \cite{Falsone2015227}, the authors prove that if $\mathcal{P}_\phi(f^*)$ is sufficiently high, then $\mathcal{I}_j$ takes positive values when $\varphi_j \in f^*$ and negative otherwise. 

\section{RFSC algorithm}
\label{sec:AlgDecr}

In view of the probabilistic reformulation of the FS problem discussed in the previous section, we here describe an iterative optimization approach that operates on the model distribution $\mathcal{P}_\phi(f)$ with the aim of maximizing
the average performance given by \eqref{eq:probForm0}. In detail, the RIPs are progressively updated based on the assessment of the importance of each regressor in terms of index $\mathcal{I}_j$, $j=1,\cdots, N_r$. Notice that an exact evaluation of $\mathcal{I}_j$ is not practically feasible, since it would require an exhaustive approach on the model space. Therefore, the expected values in \eqref{eq:Ii} are approximated with averages over a finite set of models extracted from the current model distribution. The procedure is stopped upon reaching a limit distribution.

Given the discrete nature of the $0-1$ loss function in FS problems, different models may result in the same classifier or in different classifiers of equal performance, unlike what happens in MSS in the nonlinear identification framework. Therefore, it may happen that different runs of the algorithm may provide different results.

At the beginning of each iteration, a set of models is extracted from the space of all possible model structures using the current Bernoullian distributions associated to the regressors. More in detail, for each model a value is extracted from all distributions, and only the regressors corresponding to a successful extraction are included in the model. Then, the parameters of each model are estimated and its performance evaluated according to the procedure explained in \ref{subsec:par_est} (if any redundant terms are detected, they are eliminated and the parameters re-estimated prior to evaluation). Then, the following update law is applied to the Bernoullian distribution of each regressor at the $t$th iteration:
\begin{equation}
\label{eq:update}
    \mu_j(t+1) = \text{sat}(\mu_j(t) + \gamma \tilde{\mathcal{I}}_j)
\end{equation}
for $j=1, \cdots, N_r$, where $\tilde{\mathcal{I}}_j$ is an approximation of $\mathcal{I}_j$ calculated on the set of extracted models, $\text{sat}(x) = \min(\max(x,0),1)$ is a function that ensures that the calculated $\mu_j$ values will not exceed the interval $[0,1]$, and $\gamma$ is a step-size parameter. The value of the latter parameter is adapted at each iteration:
\begin{equation}
\label{eq:gama}
    \gamma= \frac{1}{10(J_{max}-\bar{J})+0.1},
\end{equation}
where $J_{max}$ is the performance index of the best model among those extracted at the current iteration and $\bar{J}$ is their average performance. In practice, the larger the variance of the model performances, the smaller the step-size, indicating a lower level of reliability of the computed global measure of the regressor importance $\mathcal{I}_j$. 
The procedure is iterated as long as the RIPs continue to be modified.

A pseudocode of the basic loop of the proposed RFSC procedure is presented below as Algorithm  \ref{alg:alg1}.
\begin{algorithm}
	\caption{Pseudocode of the main loop of the RFSC.}
	\label{alg:alg1}
	\begin{algorithmic}[1]
	\REQUIRE{$\{ \mathbf{u}(k), \mathbf{y}(k)\}$, $\mathcal{R}= \{ \varphi_1, \ldots, \varphi_{N_r} \}$, $N_p$, $\alpha$, $\mu(0)$, $\epsilon$}
	\ENSURE{$\mu$}
	\REPEAT
		\FOR{$n_{p}=1$ \TO $N_{p}$}
			\STATE Generate random model structure $\in \mathcal{F}$\;
			\STATE Estimate parameter vector $\vartheta$ by solving \eqref{eq:minprob}\;
			\STATE Compute $\hat{\sigma}^2_j$ according to \eqref{eq:varest}\;
			\STATE Apply statistical test according to \eqref{eq:stat_tets}\;
			\STATE Remove redundant terms\;
			\STATE Re-estimate parameter vector $\vartheta$\;
			\STATE Evaluate model performance according to \eqref{eq:perfIndex}\;
		\ENDFOR
	    \FOR{ $j=1$ \TO $N_r$}
	        \STATE Evaluate importance of $j$th term using \eqref{eq:Ii}\;
	        \STATE Update $j$th RIP according to \eqref{eq:update}\;
	    \ENDFOR
        \STATE $t \gets t+1$
	\UNTIL{$\underset{j=1, \ldots, N_r}{\max} | \mu_j(t)-\mu_j(t-1)| \leq \epsilon $}
	\end{algorithmic}
\end{algorithm}

\section{RFSC with Distance Correlation Filtering}

A high-dimensional feature space can hamper FS algorithms by slowing down the search process and by increasing the chances of getting stuck in local minima. To tackle this issue a common approach is to perform a prefiltering of the feature space. Specifically, it would be desirable to identify those features that are relevant in describing the output, and those which are not. We address this problem by analyzing the dependence of the output on each feature, according to the rationale that if feature $u_p$ is not important in the description of the output $y_i$, then we would expect $y_i$ and $u_p$ to be independent. The reader should note that at this point we are just interested in characterizing the dependence/independence of the output from a specific feature, not the strength nor the ``shape'' of such dependence, tasks that are performed during the FS process.

There exist various statistical tests designed to assess the dependence between two random vectors. We here employ the one described in \cite{szekely2007}, which is based on a statistic named ``distance correlation''. The statistical test in \cite{szekely2007} is very flexible and can handle both discrete and continuous random vectors, without any assumption on their distributions, making it particularly amenable for classification purposes. For the sake of completeness, we here briefly report the main result of \cite{szekely2007}.

Let $X$ and $Y$ be two random variables such that $\mathbb{E}[|X|+|Y|]<\infty$, where $|\cdot|$ denotes the absolute value. In our case we have $X = u_p$ and $Y = y_i$ for any $i$ and $p$. We want to test the null hypothesis
\begin{equation*}
	H_0: X \text{ and } Y \text{ independent}.
\end{equation*}
Let $\mathbf{X} = [u_p(1)\,\cdots\,u_p(N)]^T$ be a vector of i.i.d. realizations of $X$, and $\mathbf{Y} = [y_i(1)\,\cdots\,y_i(N)]^T$ the corresponding realizations of $Y$. Now define the ``empirical distance covariance'' as
\begin{equation} \label{eq:dCovv}
	\nu^2_N(\mathbf{X},\mathbf{Y}) = \frac{1}{N^2} \sum_{r,s=1}^{n} A_{r,s}B_{r,s},
\end{equation}
where
\begin{align*}
	A_{rs} &= a_{rs}-\bar{a}_{r \cdot}-\bar{a}_{\cdot s} + \bar{a}_{\cdot \cdot}, \\
	B_{rs} &= b_{rs}-\bar{b}_{r \cdot}-\bar{b}_{\cdot s} + \bar{b}_{\cdot \cdot},
\end{align*}
with
\begin{equation*}
	a_{rs} = |u_p(r) - u_p(s)|, \quad b_{rs} = |y_i(r)-y_i(s)|,
\end{equation*}
and
\begin{align*}
	\bar{a}_{r \cdot} = \frac{1}{N} \sum_{s=1}^N a_{rs}, \;\;\;
	\bar{a}_{\cdot s} &= \frac{1}{N} \sum_{r=1}^N a_{rs}, \;\;\;
	\bar{a}_{\cdot \cdot} = \frac{1}{N^2} \sum_{r,s=1}^N a_{rs}, \\
	\bar{b}_{r \cdot} = \frac{1}{N} \sum_{s=1}^N b_{rs}, \;\;\;
	\bar{b}_{\cdot s} &= \frac{1}{N} \sum_{r=1}^N b_{rs}, \;\;\;
	\bar{b}_{\cdot \cdot} = \frac{1}{N^2} \sum_{r,s=1}^N b_{rs}.
\end{align*}
The interested reader is referred to \cite{szekely2007} for detailed information on $\nu_N(\mathbf{X},\mathbf{Y})$ and its properties.

The statistical test proposed in \cite{szekely2007} rejects $H_0$ if
\begin{equation} \label{eq:test_ineq}
	\frac{N \, \nu^2_N(\mathbf{X},\mathbf{Y})}{S} > \mathcal{N}^{-1}\!\left( 1-\frac{\alpha_d}{2} \right)^2,
\end{equation}
where $\mathcal{N}(\cdot)$ denotes the normal cumulative distribution function, $\alpha_d$ is the significance level of the test, and
\begin{equation} \label{eq:S_definition}
	S = \bar{a}_{\cdot \cdot} \bar{b}_{\cdot \cdot}.
\end{equation}

For each $i$, inequality \eqref{eq:test_ineq} is tested for all $p$, and only those features $u_p$ for which there is enough statistical evidence to reject the independence hypothesis are considered in the FS process for determining the classifier $\hat{y}_i$. The prefiltering procedure, denoted distance correlation filtering (DCF), is summarized in Algorithm~\ref{alg:test_procedure}.

\begin{algorithm}[b]
	\caption{Feature set preprocessing for class $i$.}
	\label{alg:test_procedure}
	\begin{algorithmic}[1]
		\REQUIRE $\{ \mathbf{u}(k), y_i(k)\}$, $\mathcal{F}_s = \{ u_1, \dots, u_{N_f} \}$, $\alpha_d$
		\ENSURE $\tilde{\mathcal{F}}^i_s$
		\STATE $\tilde{\mathcal{F}}^i_s \gets \mathcal{F}_s$
		\FOR{$j=1$ \TO $N_f$}
			\STATE $H_0^j \gets \TRUE$
			\STATE $\mathbf{X} \gets [u_j(1)\,\cdots\,u_j(N)]^T$
			\STATE $\mathbf{Y} \gets [y_i(1)\,\cdots\,y_i(N)]^T$
			\STATE Compute $\nu^2_N(\mathbf{X},\mathbf{Y})$ as in \eqref{eq:dCovv}
			\STATE Compute $S$ as in \eqref{eq:S_definition}
			\IF{ $N\nu^2_N(\mathbf{X},\mathbf{Y})/S > \mathcal{N}^{-1}\!\left( 1-\alpha_d/2 \right)^2$ }
				\STATE $H_0^j \gets \FALSE$
			\ENDIF
			\IF{ $H_0^j$ }
				\STATE $\tilde{\mathcal{F}}^i_s \gets \tilde{\mathcal{F}}^i_s \setminus \{ u_j \}$
			\ENDIF
		\ENDFOR
	\end{algorithmic}
\end{algorithm}

\section{Experimental study}
\label{sec:ExpStdy}

\subsection{Experiment design and datasets}
This section illustrates various experiments carried out to assess the performance of the proposed algorithm. Six numerical datasets from the UCI machine learning repository \cite{database} have been analyzed. The main features of the selected datasets are given in Table~\ref{tab:Tab_feat}.

\begin{table}[b]
\centering
\caption{Main characteristics of the used datasets.}
\begin{tabular}{cccccc}
Dataset & No. of  & No. of   & \multicolumn{2}{c}{Type of features} & No. of  \\
name     & samples & features & Real    & Integer   & classes \\
\hline
Bupa     & 345     & 6        & 1       & 5         & 2 \\
Ionosphere & 351   & 34       & 32      & 1         & 2 \\
Iris     & 150     & 4        & 4       & 0         & 3 \\
Sonar    & 208     & 60       & 60      & 0         & 2 \\
WDBC     & 569     & 30       & 13      & 0         & 2 \\
Wine     & 178     & 13       & 13      & 0         & 3 \\
\end{tabular}
\label{tab:Tab_feat}
\end{table}

All the input data in the original feature sets have been normalized in the range  $[0,1]$ range according to:
\begin{equation}
   u_{p}(k) = \frac{u_{p,raw}(k)-u_{p_{min}}}{u_{p_{max}}-u_{p_{min}}},
\label{eg:normalization}
\end{equation}
for $k=1,\ldots, N$, where $u_{p,raw}(k)$ is the original numeric value of the $k$th observation of feature $p$ in a given dataset, and $u_{p_{max}}$ and $u_{p_{min}}$ represent the maximum and minimum value of the $p$th attribute in the dataset, respectively.

The classification performance of the proposed algorithm on the selected datasets has been evaluated using the 10-fold cross validation (10-FCV) approach. Briefly, the dataset is split into ten (equal and non-overlapping) subsets (folds), possibly uniformly representative of all classes. Nine folds are used for training and the remaining one for testing, the procedure being repeated 10 times so that all folds are tested once. The algorithm performance is finally computed as the average over the ten independent runs. Given the randomized nature of the RFSC, different results may be obtained on each run, especially on datasets with large feature sets, for which full exploration may be too costly. For this reason, the application of the RFSC on each fold is repeated 10 times and the best model retained.

The classifier performance can be evaluated in terms of the percentage of correct classifications. In addition, we provide an alternative accuracy measure, namely the Cohen's Kappa rate \cite{ben2008comparison}, which is  capable of dealing more effectively with imbalanced data.
%
%
%
The Kappa statistic was originally designed to compare two different classifiers to measure the degree of (dis)agreement, compensating for chance (dis)agreements, but can be used to evaluate the merit of a specific classifier by comparing it to an ``ideal'' classifier producing the exact classifications. Let the \emph{confusion matrix} be an $N_c \times N_c$ matrix $C$ such that $C_{ij}$ equals the number of samples that are classified in class $i$ by classifier 1 and $j$ by classifier 2, and denote by  $C_{i\cdot} = \sum_{k=1}^{N_c} C_{ik}$ and $C_{\cdot j} = \sum_{k=1}^{N_c} C_{kj}$ the row and column counts (that represent the individual classification counts). Then, the Kappa rate is defined as follows:
\begin{equation*}
  K = \frac{N \sum_{i=1}^{N_c} C_{ii} - \sum_{i=1}^{N_c} C_{i\cdot}C_{\cdot i} }
           {N^2 - \sum_{i=1}^{N_c} C_{i\cdot}C_{\cdot i}},
\end{equation*}
and ranges from $-1$ (total disagreement) to $0$ (random classification) to $1$ (total agreement). The Kappa statistic is very useful for multi-class problems, in that it measures the classifier accuracy while compensating for random successes \cite{cano2013weighted}.

Regarding the initial parameter setup for the RFSC, the number of iterations was set to $N_i=300$, the maximum nonlinearity degree to $N_d=2$, the number of generated models to $N_p=100$, the significance confidence interval to $\alpha=0.99$ 
and all initial RIPs to $\mu_0 = 1/{N_r}$. 
Parameter $\alpha$ also influences the average model size, by acting on the threshold for the rejection of redundant terms. The closer $\alpha$ is to 1, the more regressors are rejected (and greater is the confidence that only meaningful regressors are retained), and the smaller is the average model size. 
The proposed algorithm was implemented in Matlab (version 2012b) and executed on an Intel(R) Core i7-3630QM machine, with 2.4GHz CPU, 8GB of RAM, and a 64-bit Operating System. 



\subsection{An illustration example}
To get a greater insight in the mechanisms of the selection process, we here illustrate the RFSC behavior with reference to the WDBC dataset, which has 30 attributes and 2 class labels. Assuming a maximum nonlinearity degree of $N_d=2$, the total number of extended regressors is $N_r=496$. 
We will focus on two independent runs of the RFSC algorithm. Both runs returned a 7-terms model (denoted Model 1 and Model 2) with no common regressors and only one common feature. We refer to the regressors of the returned models as ``final'' regressors. It is worth mentioning that despite their different structure, Model 1 and Model 2 both exhibit $0$ classification errors on the validation dataset.

\begin{figure}[h]
\centering
\includegraphics[width=\columnwidth]{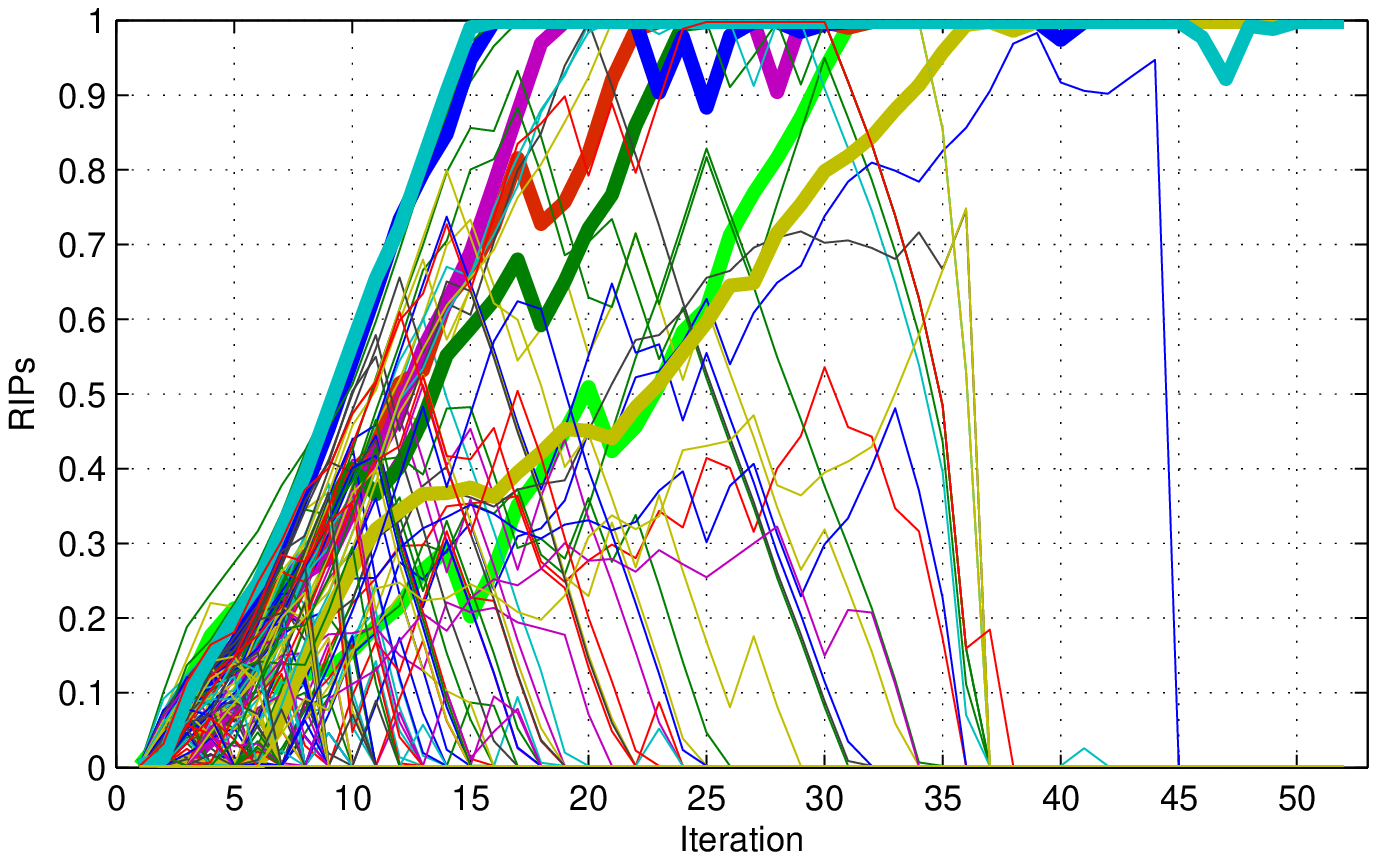}
\includegraphics[width=\columnwidth]{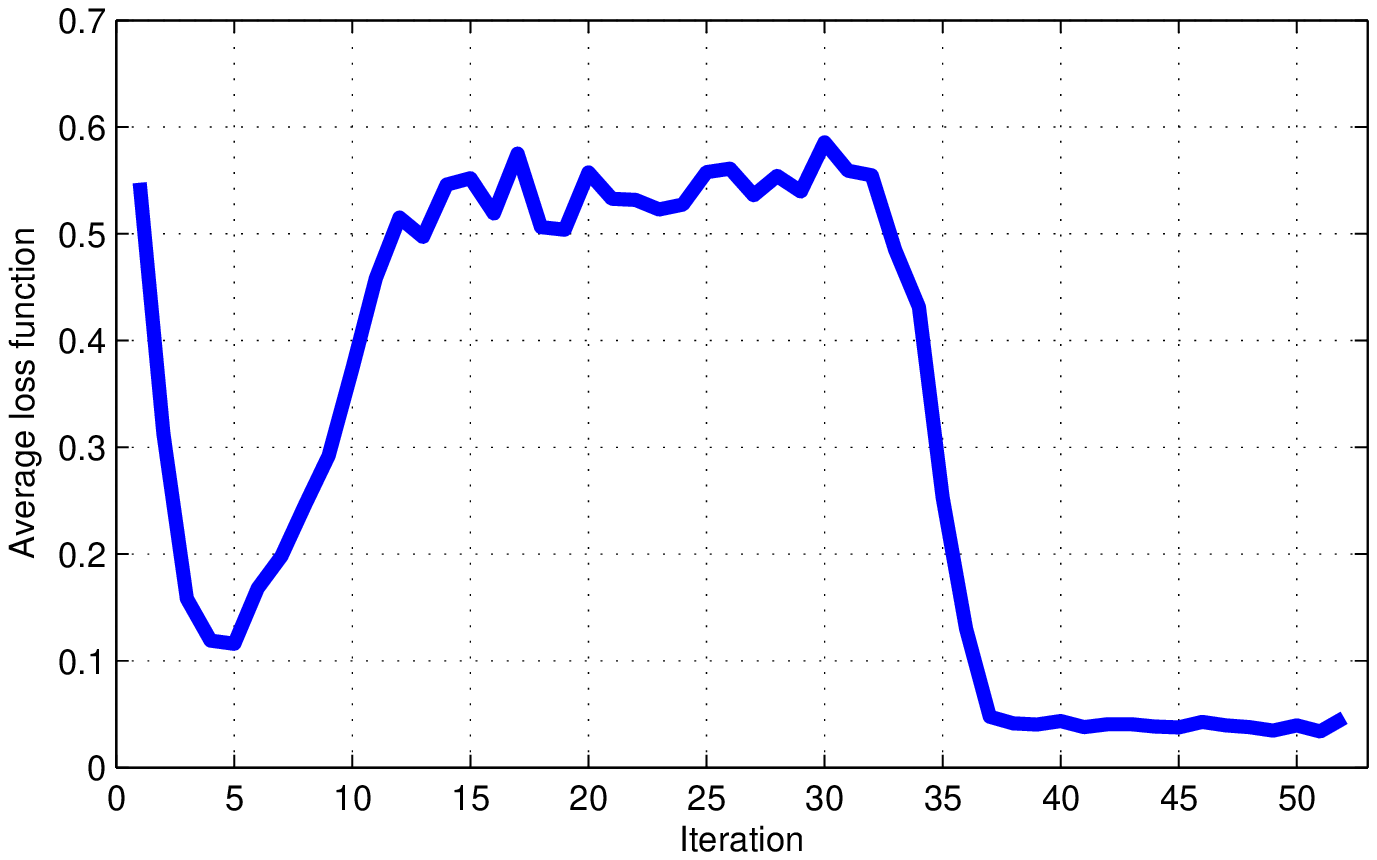}
\includegraphics[width=\columnwidth]{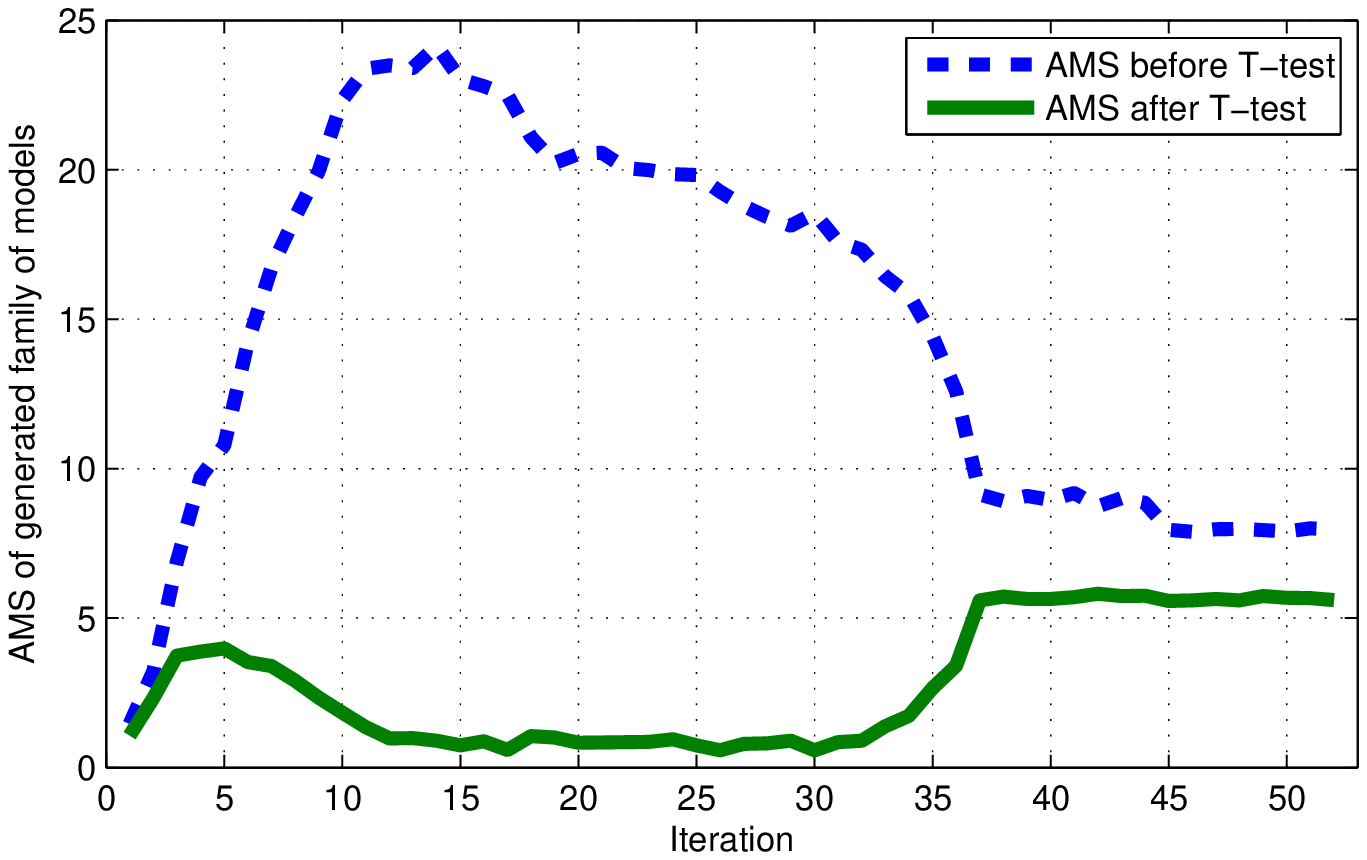}
\caption{Model 1: Evolution of the RIPs during the selection process (top, thicker lines indicate the terms contained in the final model), average loss function (middle), average model size (bottom) before (dashed) and after (solid) the t-test.}
\label{fig:Model1}
\end{figure}

\begin{figure}[h]
\centering
\includegraphics[width=\columnwidth]{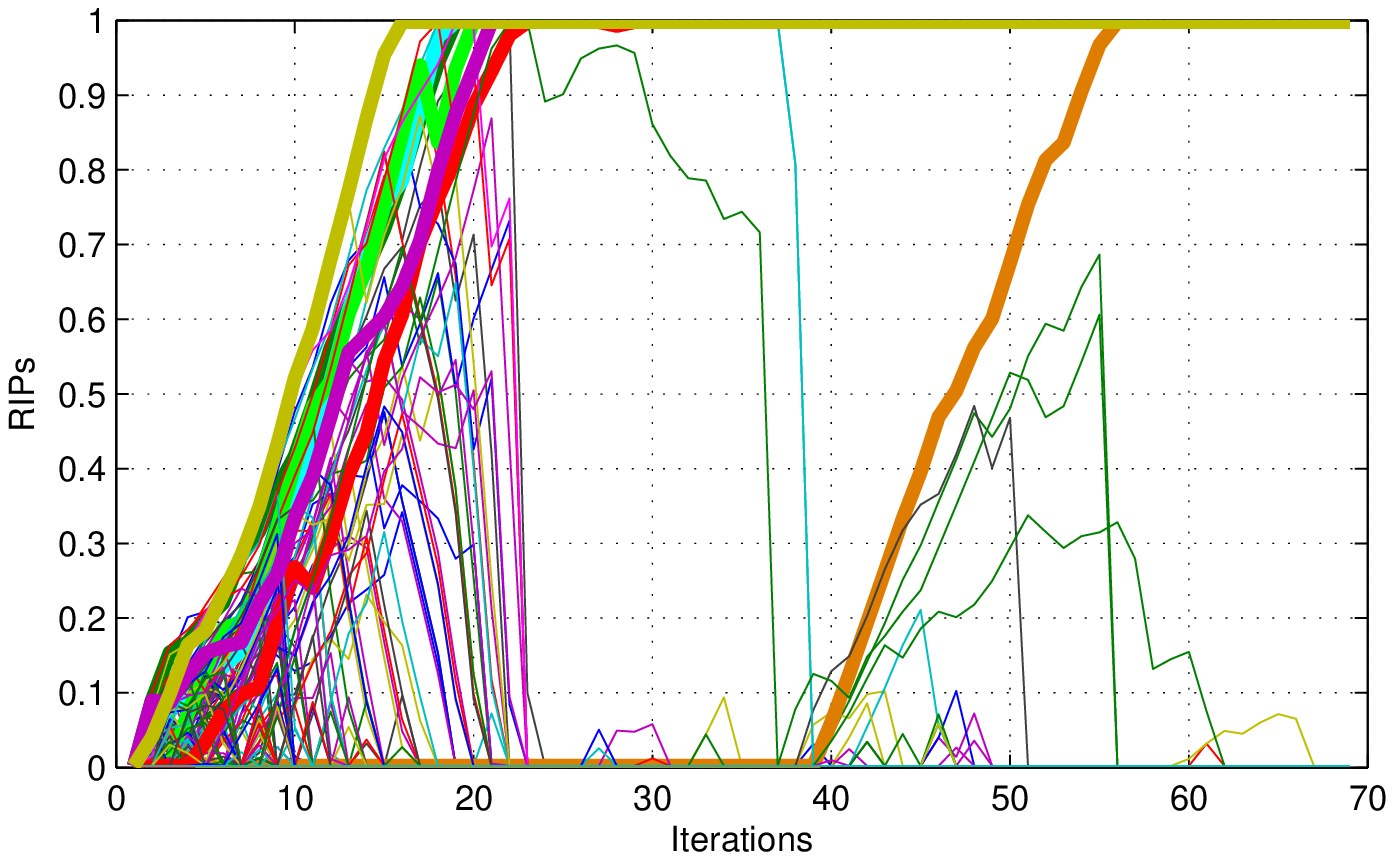}
\includegraphics[width=\columnwidth]{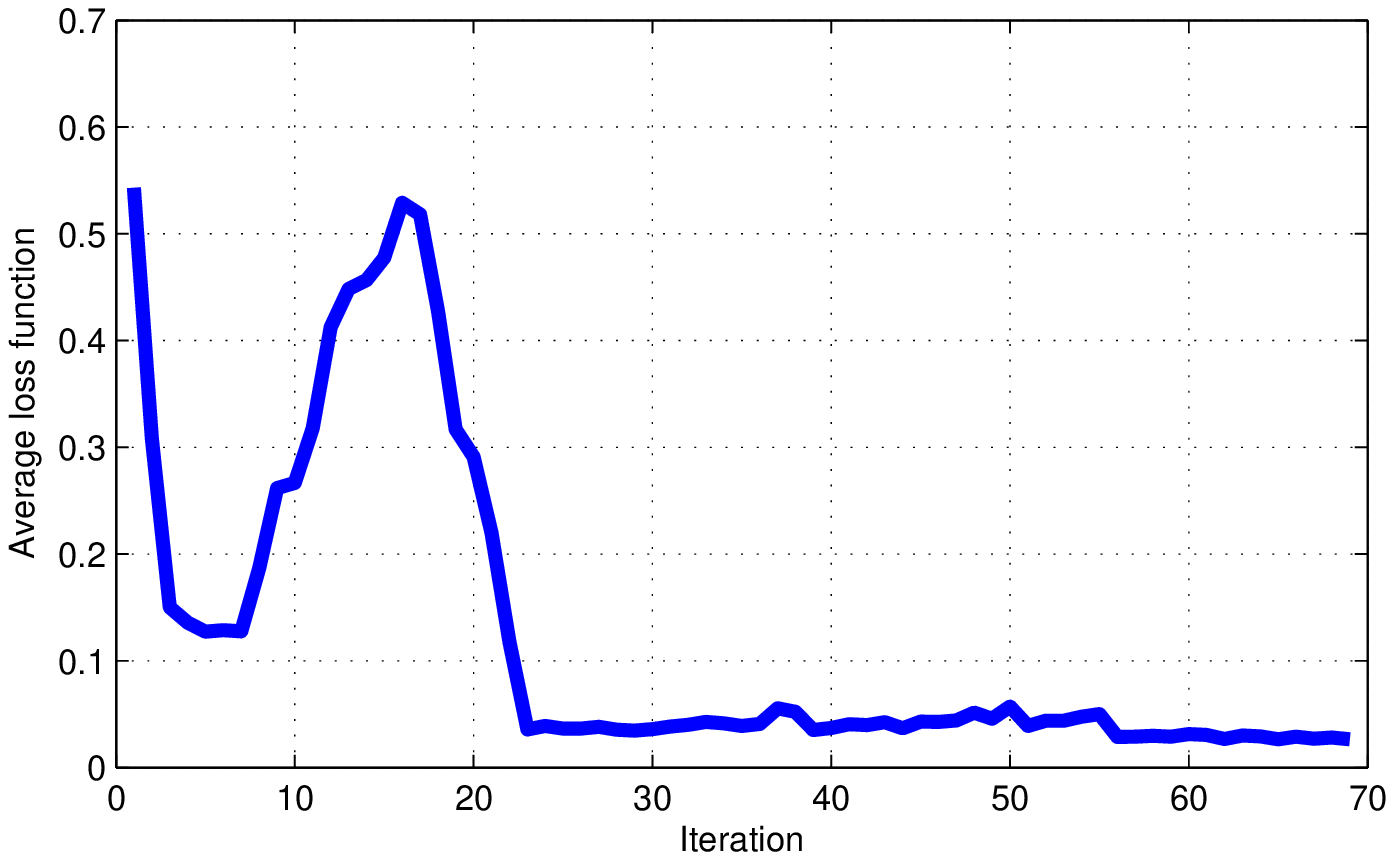}
\includegraphics[width=\columnwidth]{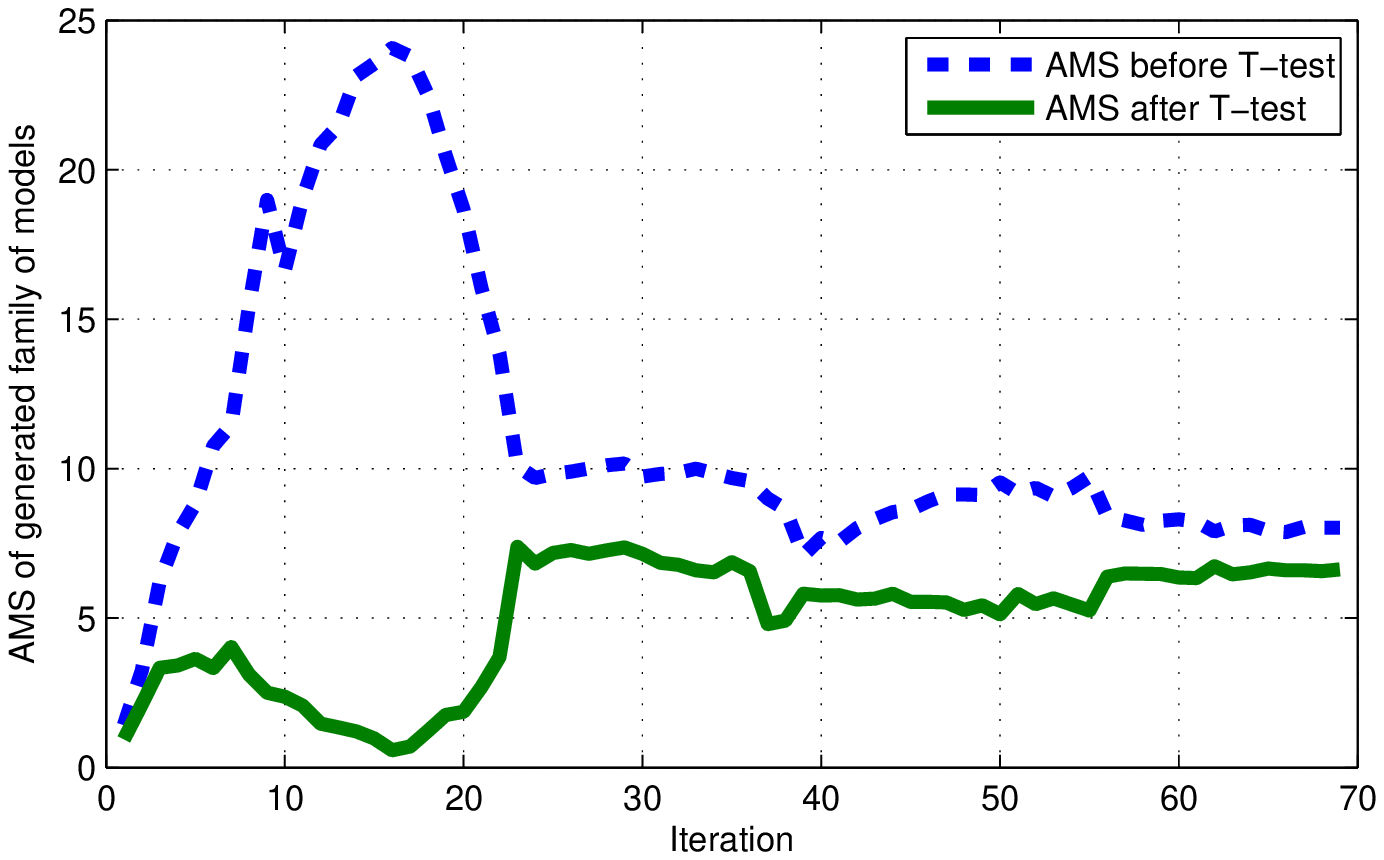}
\caption{Model 2: Evolution of the RIPs during the selection process (top, thicker lines indicate the terms contained in the final model), average loss function (middle), average model size (bottom) before (dashed) and after (solid) the t-test.}
\label{fig:Model2}
\end{figure}

Figures~\ref{fig:Model1}-\ref{fig:Model2} (top) show the evolution of the RIP values for both runs. In both cases various regressors are initially considered promising and their RIPs increased. In the first run (Fig.~\ref{fig:Model1}, top) the RIPs of the final regressors keep increasing from the very first iterations and the other regressors are progressively discarded as the algorithm progresses. On the other hand, in the second run (Fig.~\ref{fig:Model2}, top) most regressors are selected or discarded in the first $25$ iterations, but the last regressor is selected at a later stage (around iteration $40$), essentially after two other terms have been rejected. Before final convergence, other regressors are tested but ultimately discarded.
It is interesting to note that in both cases some regressors are initially selected, to the point that their RIPs rise to $1$, but are subsequently rejected in favor of other terms. If we compare (column-wise) this behavior of the RIPs with the evolution of the average loss function (average value of the loss function of the $N_p$ extracted models at a given iteration) in Figures~\ref{fig:Model1}-\ref{fig:Model2} (middle), it is clear that the algorithm is exploring model structures with a higher average loss function in order to ultimately escape from structures that represent only local minima. 


%

Figures~\ref{fig:Model1}-\ref{fig:Model2} (bottom) show the average model size (AMS) at each iteration for both runs. For Model 1, the AMS of the generated models (measured before the application of the statistical test) grows rapidly in the beginning and starts decreasing significantly only after iteration $10$. Later on, after iteration $38$, the model size does not change significantly. On the other hand, the AMS measured after the statistical test is very low from iterations $10$ to $30$, indicating that the algorithm is systematically rejecting tentative regressors as redundant. It is only between iterations $30$ to $40$ (\emph{i.e.}, when the final two regressors have been added), that the model size converges to its final value.
Similarly, for Model 2 the AMS before the t-test increases at the beginning, reaching a peak around iteration $15$, and then it stabilizes after iteration $20$. 
Notice that in both runs the AMS value is always reduced after the test, indicating the effectiveness of the latter in detecting redundant terms.

\subsection{Interpretation of the results}

As previously stated, all input data points $u_p(k)$, with $p = 1,\dots,N_f$ and $k=1,\dots,N$, have been normalized to be in the $[0,1]$ interval. Since each regressor $\varphi_j(k)$ is constructed as a product of features, $\varphi_j(k)$ takes values in $[0,1]$ as well, for all $j=1,\dots,N_r$ and $k=1,\dots,N$.

Now, the estimated model is of the form \eqref{eq:nl_reg}, where only the selected regressors are associated to non-zero parameters. The predicted class for the $k$th observation is given only by the sign of $\hat{y}_i$, while the absolute value of $\hat{y}_i$ is related to the reliability of the prediction. Since $\varphi_j(k)$ is non-negative, the information about the sign is carried by the coefficients $\vartheta^{(i)}$ of the linear combination in \eqref{eq:nl_reg}. Therefore, the model can be decomposed in two additive components based simply on the sign of the parameters:
\begin{equation}
  \label{eq:nl_reg2}
  \hat{y}_i(k) = \hat{y}_i^+(k) - \hat{y}_i^-(k) = \Phi(k)_+^T \vartheta_+^{(i)} - \Phi(k)_-^T (-\vartheta_-^{(i)}),
\end{equation}
where the first component $\hat{y}_i^+(k) = \Phi(k)_+^T \vartheta_+^{(i)}$ is associated to terms with positive coefficients and the second one $\hat{y}_i^-(k) = \Phi(k)_-^T (-\vartheta_-^{(i)})$ to terms with negative coefficients. This decomposition has the following very nice and clear interpretation: features which appear in regressors inside $\hat{y}_i^+(k)$ are representative for class $i$, whereas features appearing in $\hat{y}_i^-(k)$ are against class $i$. The ``strongest'' group of (extended) features in the $i$th model determines the sign of $\hat{y}_i$, and therefore if the predicted class should be class $i$ or not.
If multiple classes exhibit a positive $\hat{y}_i$, then the class is determined by the most ``confident'' classifier, \emph{i.e} the one with the largest difference between $\hat{y}_i^+(k)$ and $\hat{y}_i^-(k)$.

\begin{figure*}[t]
\centering
\includegraphics[width=0.85\textwidth]{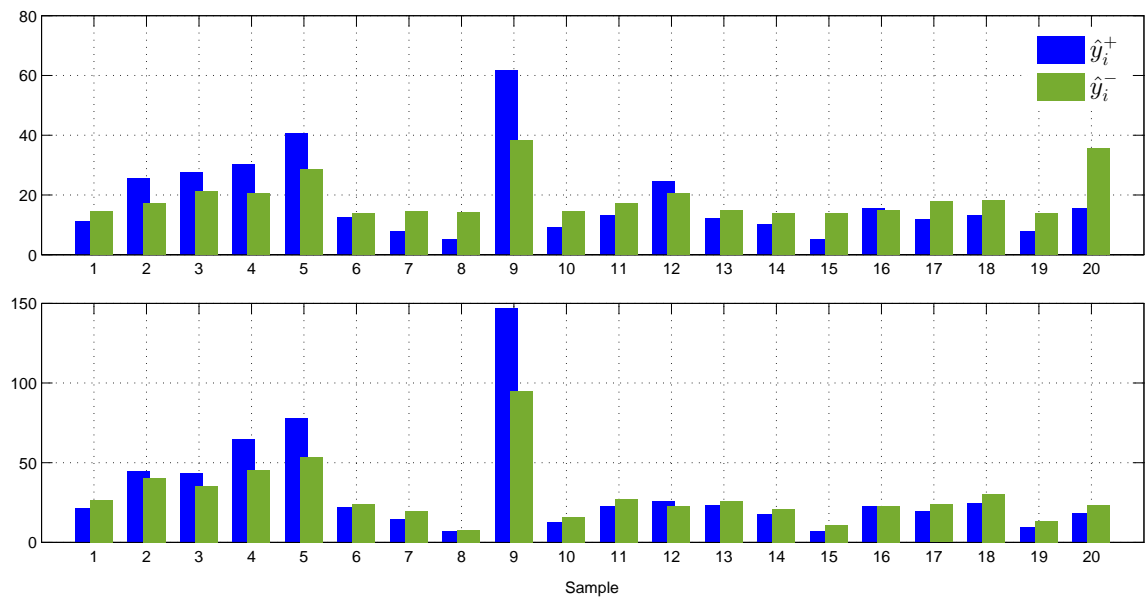}
\caption{Values of $\hat{y}_i^+(k)$ and $\hat{y}_i^-(k)$ for $20$ validation samples: Model 1 (top) and Model 2 (bottom).}%
\label{fig:contr}
\end{figure*}

In Figure~\ref{fig:contr}, we report the values of the two quantities $\hat{y}_i^+(k)$ and $\hat{y}_i^-(k)$ for $20$ validation data points. The two plots in Figure~\ref{fig:contr} (top and bottom) refer to the final models of the two runs of the RFSC algorithm analyzed in the previous section. Both models exhibit $0$ classification errors on the validation set ($56$ samples).

From Figure~\ref{fig:contr}, it is also apparent that despite the fact that both models exhibit $0$ classification errors, they are not equivalent in terms of reliability. In particular, the value of $\delta_i(k) = (\hat{y}_i^+(k) - \hat{y}_i^-(k))/\!\max (\hat{y}_i^+(k), \hat{y}_i^-(k))$ can be interpreted as the ``confidence'' the model has in attributing class $i$ to the $k$th sample. Apparently, Model 1 has generally greater values of $\delta_i$. This difference is not currently captured by the performance index \eqref{eq:perfIndex}, and therefore the two models are considered equivalent for the RFSC algorithm.

To conclude the analysis of the results, we report in Table~\ref{modelsize} the average size of the final model structures obtained by the 10-FCV procedure. Specifically, Table~\ref{modelsize} displays the number of original attributes $N_a$, the number of attributes after the DCF procedure $N_a^*$, the average number of attributes $n_a$ used by the classifier over the $10$ folds, the number of regressors $N_r$ generated based on the original attributes, the number of regressors $N_r^*$ generated based on the filtered attributes, the average number of regressors $n_r$ used by the classifier over the $10$ folds. In the non-binary classification problems (Iris and Wine datasets), a separate modeling is carried out for each class. In those cases, the classifier size (in terms of number of used features and regressors) is calculated by performing the union over the individual class models $\hat{y}_i$, $i=1, \ldots, N_c$.

By inspecting Table~\ref{modelsize}, it is noticeable that while the RFSC algorithm employs a considerable fraction of the available features, it generally requires only a small number of regressors, demonstrating its capability of compressing the information in few terms.



\begin{table}[b]
\centering
\caption{Average size of the obtained classifiers over the $10$ folds.}
\begin{tabular}{l|ccc|ccc}
Dataset     & $N_a$ & $N_a^*$ & $n_a$ &  $N_r$ & $N_r^*$ & $n_r$ \\ \hline
			&       &         &       &        &         &       \\[-1em]
Bupa        &  6    & $-$     &  5.8  &   28   & $-$     &  7.4  \\
Ionosphere  & 34    & 29      & 16.4  &  595   & 465     & 14.7  \\
Iris        &  4    & $-$     &  3.2  &   15   & $-$     &  6.1  \\
Sonar       & 60    & 39      & 25.8  & 1891   & 820     & 18.7  \\
WDBC        & 30    & 24      & 11.5  &  496   & 325     & 10.3  \\
Wine        & 13    & $-$     &  7.3  &  105   & $-$     &  7.5  \\
\end{tabular}
\label{modelsize}
\end{table}

\subsection{Comparative analysis}

To assess the performance of the RFSC algorithm in comparison with other approaches in the literature, we report in this section an extensive comparison with the results documented in \cite{xue2014particle}, \cite{cano2013weighted}, \cite{sreeja2015pattern}, \cite{lin2009psolda}, regarding the datasets in Table~\ref{tab:Tab_feat}. The comparison is carried out in terms of the average classification accuracy $J_a$, the average Kappa rate $K_a$, and the average model size. The performance comparison is summarized in Table~\ref{tab:perfComp} and the size comparison in Table~\ref{tab:sizeComp}. 
The best result within a column is highlighted. 

\begin{table*}[t]
\centering
\caption{Comparative performance analysis.}
\begin{tabular}{l|cc|cc|cc|cc|cc|cc}
FS Method + Classifier & \multicolumn{2}{c|}{Bupa} & \multicolumn{2}{c|}{Ionosphere} & \multicolumn{2}{c|}{Iris} & \multicolumn{2}{c|}{Sonar} & \multicolumn{2}{c|}{WDBC} & \multicolumn{2}{c}{Wine} \\
 & $J_a$ & $K_a$ & $J_a$ & $K_a$ & $J_a$ & $K_a$ & $J_a$ & $K_a$ & $J_a$ & $K_a$ & $J_a$ & $K_a$ \\[0.25em] \hline
 & & & & & & & & & & & \\[-0.75em]
ACO + PMC  \cite{sreeja2015pattern}              & 0.6725    & 0.3259     & \cc0.9373 & \cc0.8604  & 0.9600    & 0.9400     & \cc0.9087 & \cc0.8164  & -         & -          & 0.9755    & 0.9659     \\
Att.-Cls. WM + DGC+ \cite{cano2013weighted}      & 0.6744    & 0.3076     & 0.9311    & 0.8487     & 0.9533    & 0.9300     & 0.8487    & 0.6943     & -         & -          & 0.9731    & 0.9590     \\
Att. WV + DGC \cite{cano2013weighted}            & 0.6525    & 0.2220     & 0.6724    & 0.1142     & 0.9533    & 0.9300     & 0.7694    & 0.5187     & 0.9619    & -          & 0.9706    & 0.9552     \\
- + KNN \cite{cano2013weighted}                  & 0.6066    & 0.1944     & 0.8518    & 0.6494     & 0.9400    & 0.9100     & 0.8307    & 0.6554     & -         & -          & 0.9549    & 0.9318     \\
- + KNN-A \cite{cano2013weighted}                & 0.6257    & 0.2021     & 0.9372    & 0.8595     & 0.9533    & 0.9300     & 0.8798    & 0.7549     & -         & -          & 0.9663    & 0.9491     \\
- + DW-KNN \cite{cano2013weighted}               & 0.6376    & 0.2645     & 0.8747    & 0.7083     & 0.9400    & 0.9100     & 0.8648    & 0.7248     & -         & -          & 0.9438    & 0.9152     \\
- + Cam-NN  \cite{cano2013weighted}              & 0.5962    & 0.1024     & 0.7379    & 0.5145     & 0.9467    & 0.9200     & 0.7743    & 0.5364     & -         & -          & 0.9497    & 0.9228     \\
- + CNN  \cite{cano2013weighted}                 & 0.6316    & 0.2571     & 0.8917    & 0.7526     & 0.9267    & 0.8900     & 0.8940    & 0.7861     & -         & -          & 0.9663    & 0.9491     \\
SSMA+SFLDS  \cite{cano2013weighted}              & 0.6426    & 0.2731     & 0.9088    & 0.7986     & 0.9533    & 0.9300     & 0.8079    & 0.6100     & -         & -          & 0.9438    & 0.9145     \\
forward FS + LDA \cite{lin2009psolda}            & 0.6110    & -          & 0.8530    & -          & 0.9630    & -          & 0.7610    & -          & -         & -          & 0.9660    & -          \\
backward FS + LDA \cite{lin2009psolda}           & 0.6430    & -          & 0.9090    & -          & 0.9370    & -          & 0.8550    & -          & -         & -          & 0.9990    & -          \\
PSO + LDA  \cite{lin2009psolda}                  & 0.6520    & -          & 0.9220    & -          & \cc0.9700 & -          & 0.9050    & -          & -         & -          & \cc1.0000 & -          \\
PSO(4-2) + 5NN  \cite{xue2014particle}             & -         & -          & 0.8727    & -          & -         & -          & 0.7816    & -          & 0.9398    & -          & 0.9526    & -          \\ \hline \hline
 & & & & & \\[-1em]
(DCF) + RFSC                                     & \cc0.7800 & \cc0.4950  & 0.9330    & 0.8541     & 0.9666    & \cc0.9500  & 0.8806    & 0.8101     & \cc0.9827 & \cc0.9621  & 0.9944    & \cc0.9916  \\
\end{tabular}
\label{tab:perfComp}
\end{table*}

\begin{table}[t]
\centering
\caption{Comparative model size analysis.}
\begin{tabular}{p{6.5em}|c|c|c|c|c|c}
FS Method + Classifier & Bupa & Ion. & Iris & Sonar & WDBC & Wine \\ \hline
 & & & & & & \\[-0.75em]
FW FS + LDA          & \cc3.6 & 4.8     & \cc2.3 & \cc10.7  & -        & 7.1     \\
BW FS + LDA          & 4.7    & 30.4    & 3.9    & 56.4     & -        & 12.8    \\
PSO + LDA            & 4.6    & 21.7    & 3.6    & 38.1     & -        & 12.3    \\
PSO(4-2)+5NN         & -      & \cc3.26 & -      & 11.24    & \cc3.46  & 6.84 \\ \hline \hline
 & & & & & \\[-1em]
(DCF) + RFSC         & 5.8    & 16.4    & 3.1    & 25.8     & 11.5     & \cc3.3     \\
\end{tabular}
\label{tab:sizeComp}
\end{table}

The RFSC outperformed all other documented results on the Bupa and WDBC datasets, both in terms of average accuracy and average Kappa rate. This has been achieved at the cost of using more attributes compared to the other methods. On the other hand, the proposed algorithm was only slightly outperformed by the best competitor (which is different from case to case) on the remaining datasets, providing overall a good tradeoff between model complexity and accuracy.

\subsection{Time performance}

A comparative analysis in terms of computational time is finally presented in Table~\ref{comp_fs}. Though inherently time consuming due to model exploration mechanism in the randomized MSS process, the RFSC achieves convergence in comparable time with competitor algorithms. Indeed, it outperforms the PSO4-2 method for the Wine and WDBC datasets, but is generally somewhat slower than PSO+LDA. In this respect, it is important to note that non-optimized Matlab code has been used to obtain the documented results, so that the reported figures must be considered gross upper bounds.

\begin{table}[t] 
\centering
\caption{Computation time $[min]$.}
\label{comp_fs}
\begin{tabular}{lccc}
Dataset   & DCF + RFSC (Avrg.) & PSO4-2 & PSO + LDA \\
\hline  & & & \\[-1em]
Ionosphere & 0.95     & 1.03   &  0.46 \\
Sonar      & 1.2      & 0.54   &  0.61 \\
Wine       & 0.2      & 0.31   &  0.09 \\
WDBC       & 1.1      & 2.88   &  - \\
\end{tabular}
\end{table}

\section{Conclusions}
\label{sec:concl}

A novel method has been proposed to jointly address the FS and classifier design problems, inspired by recent results in the nonlinear model identification domain. The FS problem is reformulated as a model structure selection problem where suitable nonlinear functions of the original features are evaluated for insertion in a linear regression model. Differently from commonly adopted methods, the importance of each candidate regressors is not evaluated with reference to a specific model, but to an ensemble of models, which appears to provide a more reliable information regarding the actual significance of the term. A distribution of models is used to extract the ensemble of models and is then updated based on the aggregate information gathered from the extracted models, reinforcing the probability to extract the most promising regressors. Upon convergence a limit distribution is obtained which in practice identifies a single model structure. A distance correlation filtering (DCF) method has been occasionally found to be useful in reducing the feature set by pruning features that are independent from the model output.

The proposed method has been evaluated and compared to other well-known FS and classification algorithms obtaining quite promising and competitive results, especially in terms of the tradeoff between model complexity and classification accuracy. An important feature of the method is the easy interpretability of the obtained models, which can be used to gain more insight regarding the considered problem. Finally, the computational efficiency of the proposed method has been found to be comparable to that of competitor methods.

\bibliographystyle{ieeetr}
\bibliography{ecallBib}

\begin{thebibliography}{10}

\bibitem{dash1997feature}
M.~Dash and H.~Liu, ``Feature selection for classification,'' {\em Intelligent
  data analysis}, vol.~1, no.~1, pp.~131--156, 1997.

\bibitem{chandrashekar2014survey}
G.~Chandrashekar and F.~Sahin, ``A survey on feature selection methods,'' {\em
  Computers \& Electrical Engineering}, vol.~40, no.~1, pp.~16--28, 2014.

\bibitem{xue2014particle}
B.~Xue, M.~Zhang, and W.~N. Browne, ``Particle swarm optimisation for feature
  selection in classification: Novel initialisation and updating mechanisms,''
  {\em Applied Soft Computing}, vol.~18, pp.~261--276, 2014.

\bibitem{liu2012feature}
H.~Liu and H.~Motoda, {\em Feature selection for knowledge discovery and data
  mining}, vol.~454.
\newblock Springer Science \& Business Media, 2012.

\bibitem{ferri1994comparative}
F.~Ferri, P.~Pudil, M.~Hatef, and J.~Kittler, ``Comparative study of techniques
  for large-scale feature selection,'' {\em Pattern Recognition in Practice
  IV}, pp.~403--413, 1994.

\bibitem{korenberg1988orthogonal}
M.~Korenberg, S.~Billings, Y.~Liu, and P.~McIlroy, ``Orthogonal parameter
  estimation algorithm for non-linear stochastic systems,'' {\em International
  Journal of Control}, vol.~48, no.~1, pp.~193--210, 1988.

\bibitem{piroddi2003identification}
L.~Piroddi and W.~Spinelli, ``An identification algorithm for polynomial narx
  models based on simulation error minimization,'' {\em International Journal
  of Control}, vol.~76, no.~17, pp.~1767--1781, 2003.

\bibitem{Billings13}
S.~A. Billings, {\em Nonlinear System Identification: {NARMAX} Methods in the
  Time, Frequency, and Spatio-Temporal Domains}.
\newblock Wiley, 2013.

\bibitem{smith2005genetic}
M.~G. Smith and L.~Bull, ``Genetic programming with a genetic algorithm for
  feature construction and selection,'' {\em Genetic Programming and Evolvable
  Machines}, vol.~6, no.~3, pp.~265--281, 2005.

\bibitem{yang1998feature}
J.~Yang and V.~Honavar, ``Feature subset selection using a genetic algorithm,''
  in {\em Feature extraction, construction and selection}, pp.~117--136,
  Springer, 1998.

\bibitem{xue2013particle}
B.~Xue, M.~Zhang, and W.~N. Browne, ``Particle swarm optimization for feature
  selection in classification: A multi-objective approach,'' {\em {IEEE}
  Transactions on Cybernetics}, vol.~43, no.~6, pp.~1656--1671, 2013.

\bibitem{kabir2012new}
M.~M. Kabir, M.~Shahjahan, and K.~Murase, ``A new hybrid ant colony
  optimization algorithm for feature selection,'' {\em Expert Systems with
  Applications}, vol.~39, no.~3, pp.~3747--3763, 2012.

\bibitem{diao2012feature}
R.~Diao and Q.~Shen, ``Feature selection with harmony search,'' {\em {IEEE}
  Transactions on Systems, Man, and Cybernetics, Part B: Cybernetics}, vol.~42,
  no.~6, pp.~1509--1523, 2012.

\bibitem{paliwal2009neural}
M.~Paliwal and U.~A. Kumar, ``Neural networks and statistical techniques: A
  review of applications,'' {\em Expert systems with applications}, vol.~36,
  no.~1, pp.~2--17, 2009.

\bibitem{gunn1998support}
S.~R. Gunn {\em et~al.}, ``Support vector machines for classification and
  regression,'' {\em {ISIS} technical report}, vol.~14, 1998.

\bibitem{aha1991instance}
D.~W. Aha, D.~Kibler, and M.~K. Albert, ``Instance-based learning algorithms,''
  {\em Machine learning}, vol.~6, no.~1, pp.~37--66, 1991.

\bibitem{espejo2010survey}
P.~G. Espejo, S.~Ventura, and F.~Herrera, ``A survey on the application of
  genetic programming to classification,'' {\em {IEEE} Transactions on Systems,
  Man, and Cybernetics, Part C: Applications and Reviews}, vol.~40, no.~2,
  pp.~121--144, 2010.

\bibitem{lin2008particle}
S.-W. Lin, K.-C. Ying, S.-C. Chen, and Z.-J. Lee, ``Particle swarm optimization
  for parameter determination and feature selection of support vector
  machines,'' {\em Expert systems with applications}, vol.~35, no.~4,
  pp.~1817--1824, 2008.

\bibitem{triguero2011differential}
I.~Triguero, S.~Garc{\'\i}a, and F.~Herrera, ``Differential evolution for
  optimizing the positioning of prototypes in nearest neighbor
  classification,'' {\em Pattern Recognition}, vol.~44, no.~4, pp.~901--916,
  2011.

\bibitem{li2008nearest}
B.~Li, Y.~W. Chen, and Y.~Q. Chen, ``The nearest neighbor algorithm of local
  probability centers,'' {\em {IEEE} Transactions on Systems, Man, and
  Cybernetics, Part B: Cybernetics}, vol.~38, no.~1, pp.~141--154, 2008.

\bibitem{wang2007improving}
J.~Wang, P.~Neskovic, and L.~N. Cooper, ``Improving nearest neighbor rule with
  a simple adaptive distance measure,'' {\em Pattern Recognition Letters},
  vol.~28, no.~2, pp.~207--213, 2007.

\bibitem{dudani1976distance}
S.~A. Dudani, ``The distance-weighted k-nearest-neighbor rule,'' {\em {IEEE}
  Transactions on Systems, Man and Cybernetics}, no.~4, pp.~325--327, 1976.

\bibitem{gao2007center}
Q.-B. Gao and Z.-Z. Wang, ``Center-based nearest neighbor classifier,'' {\em
  Pattern Recognition}, vol.~40, no.~1, pp.~346--349, 2007.

\bibitem{zhou2006improving}
C.~Y. Zhou and Y.~Q. Chen, ``Improving nearest neighbor classification with cam
  weighted distance,'' {\em Pattern Recognition}, vol.~39, no.~4, pp.~635--645,
  2006.

\bibitem{cano2013weighted}
A.~Cano, A.~Zafra, and S.~Ventura, ``Weighted data gravitation classification
  for standard and imbalanced data,'' {\em {IEEE} Transactions on Cybernetics},
  vol.~43, no.~6, pp.~1672--1687, 2013.

\bibitem{peng2009data}
L.~Peng, B.~Yang, Y.~Chen, and A.~Abraham, ``Data gravitation based
  classification,'' {\em Information Sciences}, vol.~179, no.~6, pp.~809--819,
  2009.

\bibitem{Falsone2015227}
A.~Falsone, L.~Piroddi, and M.~Prandini, ``A randomized algorithm for nonlinear
  model structure selection,'' {\em Automatica}, vol.~60, pp.~227--238, 2015.

\bibitem{szekely2007}
G.~J. Sz\'{e}kely, M.~L. Rizzo, and N.~K. Bakirov, ``Measuring and testing
  dependence by correlation of distances,'' {\em Ann. Statist.}, vol.~35,
  pp.~2769--2794, 12 2007.

\bibitem{bishop2006}
C.~Bishop, {\em Pattern Recognition and Machine Learning}.
\newblock Springer, New York, 2006.

\bibitem{database}
D.~Newman, S.~Hettich, C.~L. Blake, and C.~J. Merz, ``{UCI} repository of
  machine learning databases,'' 1998.

\bibitem{ben2008comparison}
A.~Ben-David, ``Comparison of classification accuracy using {C}ohen's weighted
  kappa,'' {\em Expert Systems with Applications}, vol.~34, no.~2,
  pp.~825--832, 2008.

\bibitem{sreeja2015pattern}
N.~Sreeja and A.~Sankar, ``Pattern matching based classification using ant
  colony optimization based feature selection,'' {\em Applied Soft Computing},
  vol.~31, pp.~91--102, 2015.

\bibitem{lin2009psolda}
S.-W. Lin and S.-C. Chen, ``{PSOLDA}: A particle swarm optimization approach
  for enhancing classification accuracy rate of linear discriminant analysis,''
  {\em Applied Soft Computing}, vol.~9, no.~3, pp.~1008--1015, 2009.

\end{thebibliography}

\end{document}